\documentclass[conference]{IEEEtran}
\IEEEoverridecommandlockouts
\usepackage{cite}
\usepackage{amsmath,amssymb,amsfonts}
\usepackage{algorithmic}
\usepackage{graphicx}
\usepackage{textcomp}
\usepackage{array}
\usepackage[caption=false,font=normalsize,labelfont=sf,textfont=sf]{subfig}
\usepackage{textcomp}
\usepackage{stfloats}
\usepackage{url}
\usepackage{verbatim}
\usepackage{graphicx}
\usepackage{cite}
\usepackage{multirow}
\usepackage[table]{xcolor}
\usepackage{booktabs} 
\usepackage{amssymb}
\usepackage{pifont}

\newcommand{\gt}{\rowcolor[gray]{0.95}}
\newcommand{\rt}{\textcolor[rgb]{0.75,0.25,0.25}}

\def\BibTeX{{\rm B\kern-.05em{\sc i\kern-.025em b}\kern-.08em
    T\kern-.1667em\lower.7ex\hbox{E}\kern-.125emX}}
\begin{document}
\title{FSSUWNet: Mitigating the Fragility of Pre-trained Models with Feature Enhancement for Few-Shot Semantic Segmentation in Underwater Images}
\author{
    \IEEEauthorblockN{
        Zhuohao Li\textsuperscript{\rm 1}, 
        Zhicheng Huang\textsuperscript{\rm 1},
        Wenchao Liu\textsuperscript{\rm 1},
        Zhuxin Zhang\textsuperscript{\rm 1}, 
        Jianming Miao\textsuperscript{\rm 1,2,*}
    }
    \IEEEauthorblockA{\textsuperscript{\rm 1} School of Ocean Engineering and Technology, Sun Yat-Sen University, Zhuhai 519082, China}
    \IEEEauthorblockA{\textsuperscript{\rm 2} Southern Marine Science and Engineering Guangdong Laboratory, Zhuhai 519000, China}
     \IEEEauthorblockA{\{lizhh268, huangzhch27, liuwch8, zhangzhx76\}@mail2.sysu.edu.cn, miaojm@mail.sysu.edu.cn}
     \thanks{\textsuperscript{\rm *} Corresponding author}
     \thanks{This work is supported by the National Natural Science Foundation of China (Grant 42227901).}
}

\maketitle

\begin{abstract}
Few-Shot Semantic Segmentation (FSS), which focuses on segmenting new classes in images using only a limited number of annotated examples, has recently progressed in data-scarce domains.
However, in this work, we show that the existing FSS methods often struggle to generalize to underwater environments.
Specifically, the prior features extracted by pre-trained models used as feature extractors are fragile due to the unique challenges of underwater images.
%
To address this, we propose FSSUWNet, 
a tailored FSS framework for underwater images with feature enhancement. FSSUWNet exploits the integration of complementary features, emphasizing both low-level and high-level image characteristics.
In addition to employing a pre-trained model as the primary encoder, we propose an auxiliary encoder called Feature Enhanced Encoder which extracts complementary features to better adapt to underwater scene characteristics.
Furthermore, a simple and effective Feature Alignment Module aims to provide global prior knowledge and align low-level features with high-level features in dimensions.
Given the scarcity of underwater images, we introduce a cross-validation dataset version based on the Segmentation of Underwater Imagery dataset.
Extensive experiments on public underwater segmentation datasets demonstrate that our approach achieves state-of-the-art performance. 
For example, our method outperforms the previous best method by 2.8\% and 2.6\% in terms of the mean Intersection over Union metric for 1-shot and 5-shot scenarios in the datasets, respectively.
Our implementation is available at \url{https://github.com/lizhh268/FSSUWNet}.
\end{abstract}

\section{Introduction}

In recent years, image semantic segmentation technology has become a popular research topic in various fields due to its ability to provide pixel-level target information~\cite{long2015fully, ronneberger2015u}. 
However, in real-world applications, obtaining perfectly annotated image datasets is difficult, particularly in specific domains such as underwater environments, where data scarcity has become a major challenge to research progress.
\begin{figure}[!t]
\centering
\includegraphics[width=0.45\textwidth]{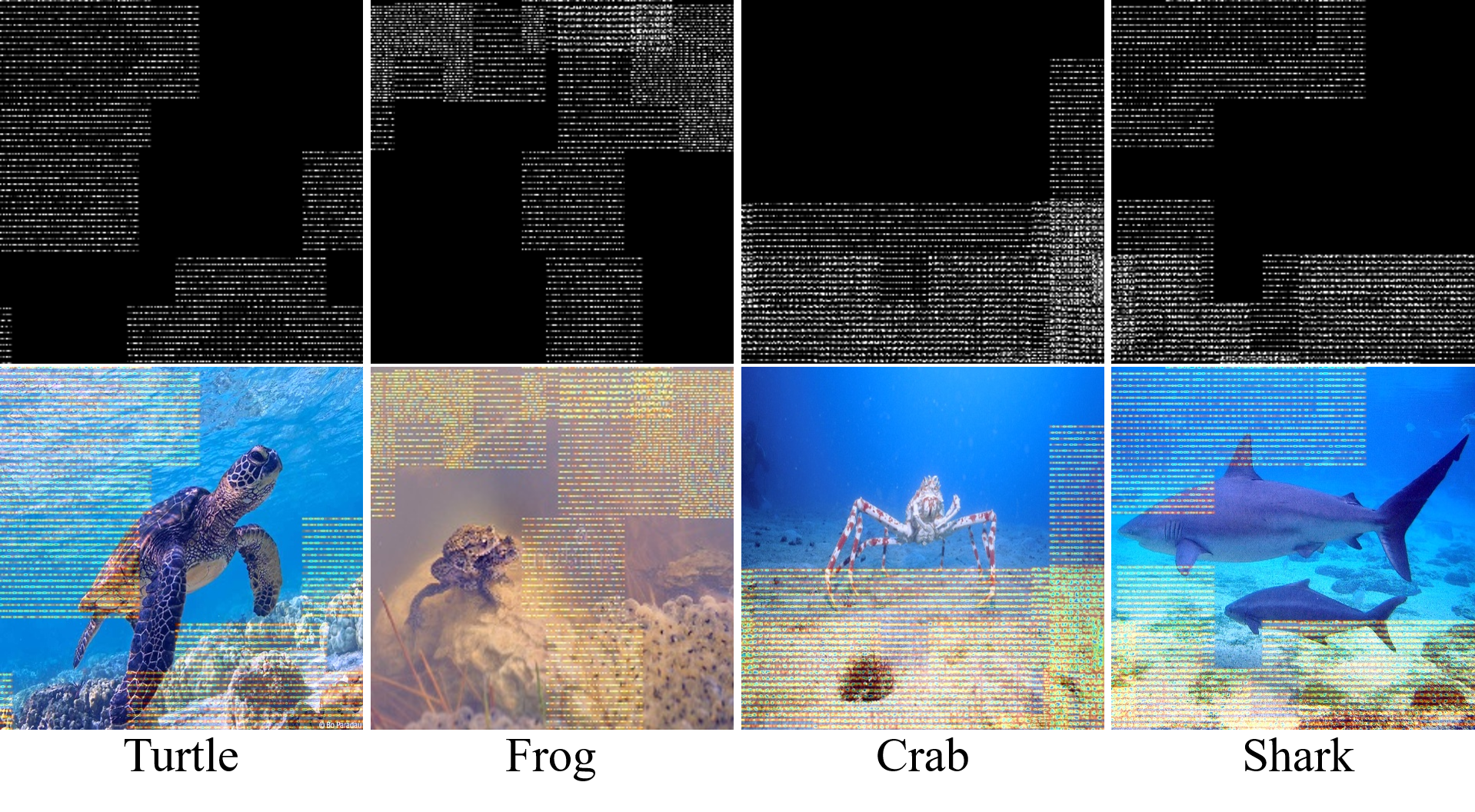}
\caption{
The fragile prior masks extracted from the pre-trained model (VGG-16) for real-world underwater images.
\textit{Top:} the prior masks of underwater images. \textit{Bottom:} Visualization of the segmentation results of the prior masks.
Existing pre-trained models fail to effectively extract the foreground regions (underwater animals) of underwater images and mistakenly identify many background areas, showing significant fragility.
}
\label{fig_prior}
\vspace{-1em}
\end{figure}
To address this issue, Few-Shot Learning (FSL) has been proposed, revolutionizing the paradigm of image recognition~\cite{vinyals2016matching, wang2016learning}.
Few-Shot Semantic Segmentation (FSS)~\cite{FSS} has demonstrated potential in semantic segmentation, enabling models to learn new objects or classes with just a few labeled images.
Nevertheless, despite its advancements, FSS technology in the underwater domain remains limited, facing more challenges than general images.
Specifically, the unique challenges posed by underwater images include:
\ding{182} Due to the water absorption effects, underwater images suffer from color bias and a loss of detail, resulting in degraded image quality~\cite{Waternet};
\ding{183} It is extremely difficult to obtain large-scale, reliable, and available underwater image data~\cite{Waternet,underwater2023iros}; 
\ding{184} There are significant differences between underwater and terrestrial environments~\cite{islam2020suim}.

Existing state-of-the-art FSS methods~\cite{panet, pfenet, lang2023bam, luo2023pfenet} rely on pre-trained models (deep convolutional networks), \emph{e.g.}, VGG~\cite{vgg} and ResNet~\cite{resnet}, to extract prior features from images.
These methods have shown promise for non-underwater scenes.
However, there is no FSS method specifically developed for underwater environments.
Moreover, in this work, we find that these approaches are not well-suited for underwater images and often fail to distinguish query objects effectively.

As shown in Figure~\ref{fig_prior}, 
we extracted the prior features and conducted visualization processing, which can be observed that the prior masks are unable to effectively cover the foreground regions (underwater animals) to be queried, and even mistakenly identified the background as the target.
%
In this work, we present that \emph{these prior features provided by pre-trained models could be fragile when applied to underwater images, which stems from the unique challenges brought by underwater environments.}
Unfortunately, 
there are no effective FSS solutions specifically tailored to these underwater image challenges, especially the fragility of pre-trained models.
To address these challenges, this work aims to mitigate the fragility of pre-trained models in existing FSS methods in underwater images.
We propose a novel few-shot semantic segmentation framework with feature enhancement, FSSUWNet, specifically designed for underwater images.
Our approach strategically integrates features extracted from pre-trained encoders and an additional auxiliary encoder, Feature Enhanced Encoder, which extracts complementary features from underwater images.
Additionally, we present a simple Feature Alignment Module to enhance the low-level features of the input images.
The enhanced features will be aligned with high-level features in scale and serve as shared global prior knowledge for both query and support images, improving the model's performance in underwater segmentation.
Furthermore, due to the limited availability of underwater images, we propose SUIM-FSS, a variant of the SUIM underwater image dataset~\cite{islam2020suim} designed for cross-validation.
Comprehensive experiments conducted on publicly available underwater image segmentation datasets (UWS dataset~\cite{underwater2023iros} and SUIM-FSS dataset) show that our proposed FSSUWNet achieves superior performance compared to existing state-of-the-art methods.
We also verified the efficacy of our proposed components and analyzed the roles of different feature levels, providing insights into feature utilization for future underwater FSS work.

In summary, our contributions are as follows:
\begin{itemize}
\item[$\bullet$] 
We show that the image prior features provided by pre-trained models in most FSS methods could be fragile in underwater scenes, which will be challenging to apply to underwater images.

\item[$\bullet$] 
We propose a novel FSS framework tailored to underwater images with the careful utilization of both low-level and high-level image features, dubbed FSSUWNet.
%
In FSSUWNet, we introduce a Feature Enhanced Encoder to adapt to underwater scene characteristics and a Feature Alignment Module to enhance and align low-level features with high-level ones.

\item[$\bullet$] 
We introduce SUIM-FSS, a variant of the SUIM underwater image dataset for cross-validation.
Extensive experiments on public underwater segmentation datasets show that the proposed FSSUWNet achieves state-of-the-art performance.

\end{itemize}
\section{Related Work}
\subsection{Semantic Segmentation}
Image semantic segmentation involves segmenting object pixels from an image and is widely applied in fields like autonomous driving~\cite{feng2020deep} and object recognition~\cite{huang2021fapn,long2015fully,ronneberger2015u}. Early approaches used CNNs as backbones with segmentation heads, such as FCNs~\cite{fcns}, which replaced fully connected layers with convolutional layers for end-to-end learning and dense pixel-wise predictions. U-Net~\cite{ronneberger2015u} added an encoder-decoder structure for capturing semantic features and performing segmentation.
Recently, Transformers~\cite{vaswani2017attention} have gained popularity in computer vision~\cite{pengUTransformerUIE2021,strudel2021segmenter,xie2021segformer,ShadowMaskFormer}, offering advantages in modeling long-range pixel dependencies. For example, Segformer~\cite{xie2021segformer} combines a Transformer encoder with a convolutional decoder, efficiently capturing global context while preserving spatial details. 
However, the success of semantic segmentation also hinges on large-scale annotated datasets. In underwater environments, acquiring such datasets remains a significant challenge due to their unique characteristics~\cite{underwater2023iros}.

\subsection{Few-Shot Learning}
Few-Shot Learning (FSL) has become a key area of research due to its ability to generalize to new tasks, such as object detection and segmentation in complex scenes~\cite{snell2017prototypical,zhang2018metagan,sun2019meta}. The meta-learning framework is commonly used in FSL, leveraging learned meta-data to handle new learning tasks. FSL can be categorized into three approaches: optimization-based methods to accelerate solution exploration, data augmentation for performance improvement, and metric-based methods, which are relevant to our work. Metric-based methods use distance metrics, such as cosine similarity, to compute the distance between support and query features.
Recent developments in metric-based methods~\cite{panet,pfenet,apanet} focus on minimizing the distance between prototypes and foreground features in the query, while maximizing the distance to background features. Our approach enhances the feature representation capability of the FSL framework for underwater scenes by introducing additional auxiliary features, ensuring more accurate distance calculations between support and query features.

\subsection{Few-Shot Semantic Segmentation}
Few-Shot Semantic Segmentation (FSS) addresses data scarcity by leveraging a small number of labeled samples, combining techniques like transfer learning and meta-learning to quickly recognize unknown objects, such as bicycles or airplanes~\cite{dong2018few,panet,yang2020prototype,pfenet}. Given its ability to work with minimal labeled data, FSS has gained attention in fields like image classification and segmentation. Recent approaches, such as BAM~\cite{lang2023bam}, use an additional base learner to predict base class regions and suppress distractor objects in query images. APANet enhances classification by differentiating prototypes into class-specific and class-agnostic categories~\cite{apanet}.
However, existing FSS methods, designed primarily for indoor and outdoor environments, struggle with generalization to underwater scenes. To address this gap, UWSNet introduced the UWS Dataset, the most comprehensive underwater FSS dataset~\cite{underwater2023iros}, but it did not tackle the limitations of single pre-trained models, like VGG~\cite{vgg}, in underwater settings. Inspired by BAM's additional base class encoder, we propose enhancing extracted features to better adapt to underwater scene characteristics through a complementary feature enhancement encoder. Additionally, recognizing the varying roles of low-level and high-level features~\cite{CANet,pfenet}, we treat these features differently to improve the generalization of prototype features across scenes.
\section{Method}
\begin{figure*}[ht]
\centering
\includegraphics[width=0.8\textwidth]{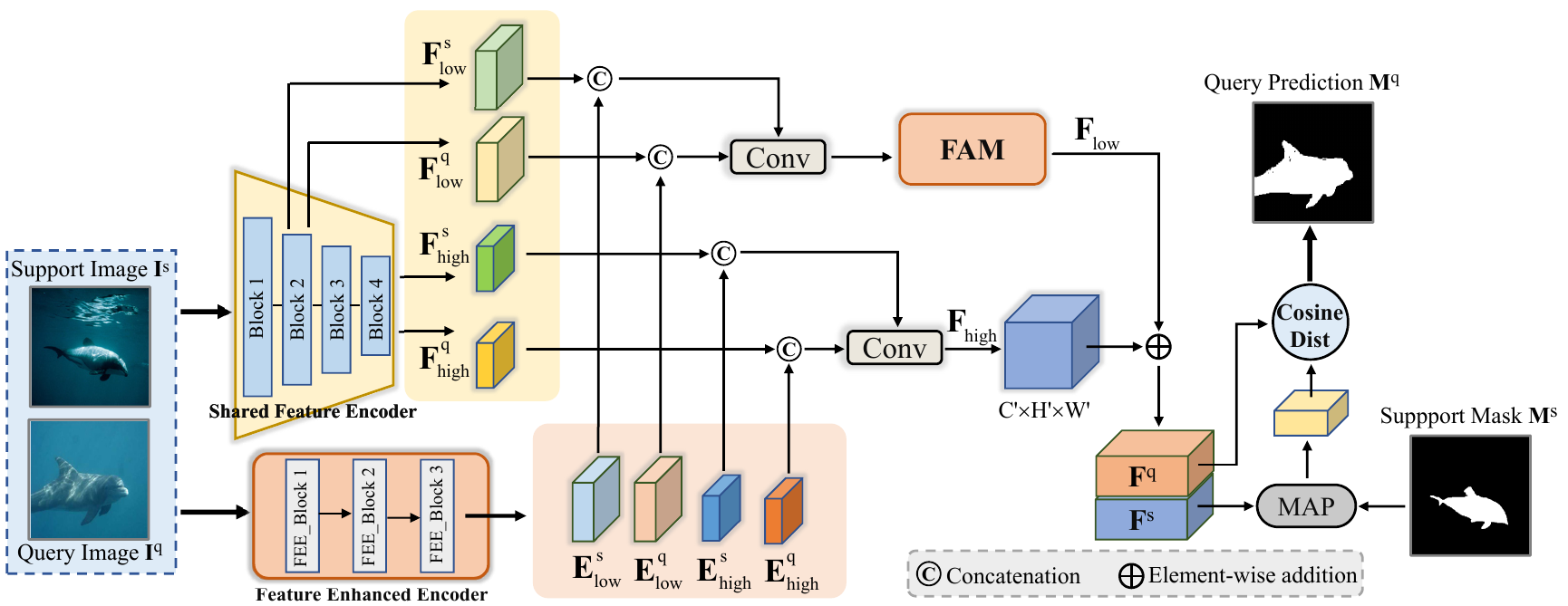}
\caption{The network architecture of the FSSUWNet with Shared Feature Encoder (SFE), Feature Enhanced Encoder (FEE), and Feature Alignment Module (FAM).
Each encoder extracts low-level and high-level features from both query and support images.
The low-level features of images are then enhanced through FEM to obtain underwater image enhancement features $\in \mathbb{R}^{C^{'} \times H^{'} \times W^{'}}$.
Subsequently, the high-level features of the images are pixel-wise addition with the enhanced image features to obtain the final query feature $\mathit{\mathbf{F}^q}$ and support feature $\mathit{\mathbf{F}^s}$.
Finally, a masked average pooling operation is conducted, followed by the computation of cosine similarity between $\mathit{\mathbf{F}^q}$ and $\mathit{\mathbf{F}^s}$.
Note that Block and FEE\_Block represent the feature extraction blocks in SFE and FEE, respectively.}
\label{fig_framework}
\end{figure*}
We will first present the essential preliminaries in Section~\ref{preliminary}, including the task definition and motivation, followed by a detailed introduction to FSSUWNet.
\subsection{Preliminary\label{preliminary}}

\subsubsection{Task description\label{task}}
Few-Shot Semantic Segmentation (FSS) addresses the challenge of segmenting novel object classes in images when the number of annotated examples is limited. 
The FSS task involves two main sets: the support set $\mathit{S}$ and the query set $\mathit{Q}$.
In each episode, the support set $\mathit{S}$ comprises $\mathit{K}$ image-mask pairs, denoted as $\mathit{S} = \{(I^s_{i}, M^s_{i})\}^{k}_{i=1}$, where each pair consists of a support image $\mathit{I^s_{i}}$ and its corresponding segmentation mask $\mathit{M^s_{i}}$ for a specific class $\mathit{c}$. The query set $\mathit{Q}$ includes a query image $\mathit{I^q}$ paired with its respective ground truth mask $\mathit{M^q}$ which is hidden during inference and used only for training.
Models are trained on a set of base classes $\mathit{C_{train}}$ and evaluated on a distinct set of unseen classes $\mathit{C_{test}}$, ensuring no overlap between them ($\mathit{C_{train}} \cap \mathit{C_{test}} = \mathit{\varnothing}$)
The combined input batch $\{\mathit{I^q}, \{(I^s_{i}, M^s_{i})\}^{k}_{i=1}\}$ provides the context necessary for the model to infer the segmentation result for the query image.

\subsubsection{Motivation\label{motivation}}
Our motivation stems from two aspects: the fragility of prior features in underwater scenes and the effective utilization of features across different levels.
\noindent\textit{\textbf{Fragility of Prior Features in Underwater Scenes.}}  
High-level semantic features, often from deeper neural network layers, are commonly used in prior works~\cite{pfenet,lang2022bam,lang2023bam} as feature encoders for query and support images. These features, along with support image masks, generate prior features and masks for query images, known as Prior Guided Features~\cite{pfenet}. However, as shown in the top row of Figure~\ref{fig_prior}, prior masks derived from BAM~\cite{lang2023bam} struggle to accurately segment underwater scenes, often misclassifying background as foreground. This fragility arises from the limited presence of underwater scenes in training datasets like ImageNet~\cite{ImageNet}, causing poor generalization. Our work addresses this by enhancing feature representations to mitigate this fragility.

\noindent\textit{\textbf{Differences Across Feature Levels.}} 
Image features vary across levels in their representational capacity. Lower-level features capture basic image characteristics, while higher-level features offer stronger semantic representations. Empirical results from CANet~\cite{CANet} show that using middle-level features from pre-trained models like VGG~\cite{vgg} and ResNet~\cite{resnet} leads to better FSS performance than using high-level features. Inspired by this, existing FSS approaches~\cite{pfenet,lang2023bam} integrate features from different levels. Our FSSUWNet framework combines low-level and high-level features to enhance segmentation performance, as detailed in Section~\ref{levels}.

\subsection{Overview of FSSUWNet\label{framework}}
The complete framework of our proposed FSSUWNet under the 1-shot setting is illustrated in Figure~\ref{fig_framework}.
Due to the color distortion caused by light absorption effects, such as the significant reduction in the red channel, we introduced our approach with feature enhancement to extract and enhance the common fundamental features of underwater images.
Our FSSUWNet consists of two encoders: Shared Feature Encoder and Feature Enhanced Encoder. These encoders extract both low-level and high-level features from the support image and query image.
In feature enhancement, we first use an auxiliary encoder to provide complementary features of underwater scenes and then design a Feature Alignment Module to align the low-level features with high-level features in dimensions.

\subsection{Complementary High-level Features from Encoders}
As shown in Figure~\ref{fig_framework}, in the 1-shot scene, similar to advanced FSS work, we use a Shared Feature Encoder to extract features from both support image $\mathit{\text{I}^s}$  and query images $\mathit{\text{I}^q}$, $\mathit{\text{I}^s}$ and $\mathit{\text{I}^q} \in \mathbb{R}^{ 3 \times H \times W}$, often using a single pre-trained model for this purpose.
Then, the high-level features $\mathit{\mathbf{F}^s_{high}} \in \mathbb{R}^{ N^{s}_f \times H^{'} \times W^{'}}$ and $\mathit{\mathbf{F}^q_{high}} \in \mathbb{R}^{ N^{q}_f \times H^{'} \times W^{'}}$, which highly represent semantic information and could be regarded as prior features.
The superscript “s” of $N^{s}_f$ stands for support, and the subscript “f” stands for Shared Feature Encoder. 
In our work, we follow~\cite{lang2022bam,underwater2023iros} to extract high features from deep blocks (Block 4 and FEE\_Block 3) in encoders.
Given the fragility when prior features meet underwater scenes as discussed in Section~\ref{motivation}, we attempt to introduce an auxiliary encoder, the Feature Enhanced Encoder (FEE), which can extract underwater scene features that prior features cannot provide. By fusing the features from both encoders, we aim to enrich the semantic features obtained from the image.
Therefore, we can further obtain high-level features $\mathit{\mathbf{E}^s_{high}} \in \mathbb{R}^{ N^{s}_e \times H^{'} \times W^{'}}$ and $\mathit{\mathbf{E}^q_{high}} \in \mathbb{R}^{ N^{q}_e \times H^{'} \times W^{'}}$ extracted from both $\mathit{\text{I}^s}$ and $\mathit{\text{I}^q}$ by FEE, respectively.
The subscript “e” of $N^{s}_e$  stands for FEE.
We concatenate the features from the two encoders along the channel dimension, and the operations can be formulated as:
\begin{equation}
\begin{aligned}
    \label{eq:1}
    \mathbf{H}_{s} = Cat(\mathbf{F}^s_{high},\  \mathbf{E}^s_{high}) \\
    \mathbf{H}_{q} = Cat(\mathbf{F}^q_{high},\  \mathbf{E}^q_{high})
\end{aligned}
\end{equation}
where $Cat$ is the concatenation operation, $\mathbf{H}_{s} \in \mathbb{R}^{ (N^{s}_f + N^{s}_e) \times H^{'} \times W^{'}}$ and $\mathbf{H}_{q} \in \mathbb{R}^{ (N^{q}_f + N^{q}_e) \times H^{'} \times W^{'}}$ represent the concatenated high-level semantic features extracted from the support image and the query image, respectively. 
Note that we concatenate $\mathbf{H}_{s}$ and $\mathbf{H}_{q}$ along the batch-size dimension, then use a 1x1 convolution operation to reduce the number of channels.
This results in the high-level features, $\mathbf{F}_{high} \in \mathbb{R}^{2 \times C^{'} \times H^{'} \times W^{'}}$.
These operations can be formulated as:
\begin{equation}
    \label{eq:2}
    \mathbf{F}_{high} = Conv (Cat(\mathbf{H}_{s},\  \mathbf{H}_{q})) ,
\end{equation}
where $Conv$ is the 1x1 convolution operation.

Next, $\mathbf{F}_{high}$ will undergo a pixel-wise addition operation with the output of the proposed FAM, $\mathbf{F}_{low} \in \mathbb{R}^{ C^{'} \times H^{'} \times W^{'}}$, which handles the low-level features from two encoders.
we can further separate the final support feature $\mathbf{F}_{s} \in \mathbb{R}^{ C^{'} \times H^{'} \times W^{'}}$ and query feature $\mathbf{F}_{q} \in \mathbb{R}^{ C^{'} \times H^{'} \times W^{'}}$, with both features incorporating low-level features $\mathbf{F}_{low}$ and high-level features $\mathbf{F}_{high}$.
We can express it as:
\begin{equation}
    \label{eq:3}
    \mathbf{F}_{s},\  \mathbf{F}_{q} = Split (\mathbf{F}_{high} \oplus \mathbf{F}_{low}) ,
\end{equation}
where $Split$ represents the separation operation implemented using dimensional slicing and $\oplus$ is the element-wise addition.
Furthermore, We construct a compact and useful prototype from the support feature $\mathbf{F}_{s}$ that represents the key features of the target object for segmentation in the query image.
A straightforward approach is to apply a global average pooling or max-pooling operation to aggregate the information contained in the support feature $\mathbf{F}_{s}$.
Following~\cite{pfenet,lang2022bam,lang2023bam,underwater2023iros}, we calculate the prototype from $\mathbf{F}_{s}$ through the masked average pooling (MAP)~\cite{zhang2020sg}:
\begin{equation}
    \label{eq:4}
    \mathbf{P}_{s} = {F}_{pool}(\mathbf{F}_{s}\ \otimes \  \mathbf{M}_s) ,
\end{equation}
where ${F}_{pool}$ denotes the average-pooling operation, while $\otimes$ 
represents Hadamard product. $\mathbf{M}_s$ is the ground truth support mask.
The symbol “$\mathbf{P}$” of $\mathbf{P}_{s}$ stands for prototype, while the symbol “$\mathbf{M}$” of $\mathbf{M}_{s}$ stands for mask.
The support prototype $\mathbf{P}_{s}$ will be a vector of dimension $1 \times C^{'}$ and $ C^{'}$ represents the channels in the query feature $\mathbf{F}_{q}$.
It should be noted that in the 5-shot setting, we create a unified prototype generated by averaging the five individual prototypes.
We then use the cosine similarity measure to compute the cosine distance $Dist$ between the support prototype $\mathbf{P}_{s}$ and the query feature $\mathbf{F}_{q}$, which can be summarized as:
\begin{equation}
    \label{eq:5}
    Dist = Cosine(\mathbf{P}_{s}\ , \mathbf{F}_q) ,
\end{equation}
where $Cosine$ represents the function of cosine similarity measurement.
Finally, the predicted mask $\mathbf{M}_{q}$ for the query image is generated through the cosine distance.

\begin{figure}[!t]
\centering
\includegraphics[width=0.45\textwidth]{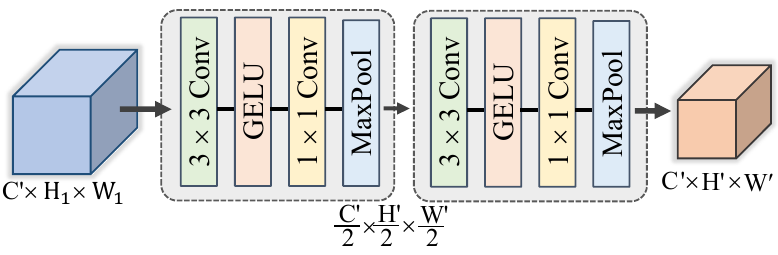}
\caption{
The network architecture of the Feature Alignment Module (FAM) in our proposed FSSUWNet involves concatenating the low-level features extracted by two encoders at the channel level and then inputting them into FAM.
In FAM, the output $\mathbf{F}_{low}$ is obtained through simple feature channel-wise operations to align with high-level features.
}
\label{fig_fem}
\end{figure}

\subsection{Feature Alignment Module}
Due to the optical absorption effect, underwater scene images exhibit unique color distortion, resulting in both support image $\mathit{\text{I}^s}$ and query images $\mathit{\text{I}^q}$ sharing certain common fundamental features. 
In FSSUWNet, we represent these common underwater image features using the low-level features extracted by the two encoders, i.e., Shared Feature Encoder and Feature Enhanced Encoder.
Similar to the high-level features ($\mathbf{F}^s_{high}, \mathbf{F}^q_{high},\mathbf{E}^s_{high},\mathbf{E}^q_{high}$), we can obtain the low-level features as follows:
\begin{equation}
\begin{aligned}
    \label{eq:low} 
    \mathbf{F}^s_{low} \in \mathbb{R}^{ C^{s}_f \times H_{1} \times W_{1}}, \ \mathbf{F}^q_{low} \in \mathbb{R}^{ C^{q}_f \times H_{1} \times W_{1}}, \\
    \mathbf{E}^s_{low} \in \mathbb{R}^{ C^{s}_e \times H_{1} \times W_{1}}, \ \mathbf{E}^q_{low} \in \mathbb{R}^{ C^{q}_e \times H_{1} \times W_{1}}, \notag
\end{aligned}
\end{equation}
where the subscript “$low$” stands for the low-level features. Note that in our work, we follow~\cite{pfenet} to obtain low-level features from shallow blocks (Block 2 and FEE\_Block 1) in encoders.

We introduced the Feature Alignment Module (FAM) to align the common fundamental features of underwater images with high-level features.
FAM is a simple but effective module that includes convolution and pooling operations. We concatenate the low-level features from the $\mathit{\text{I}^s}$ and $\mathit{\text{I}^q}$ along the channel dimension. We then use a 1x1 convolution to reduce the number of channels to simplify the network, resulting in the raw low-level features $\mathbf{F}^{0}_{low} \in \mathbb{R}^{2 \times C^{'} \times H^{'} \times W^{'}}$ which include both the support and query vectors in the batch size dimension.
These operations can be formulated as:
\begin{equation}
\small
    \label{eq:2}
    \mathbf{F}^{0}_{low} = Conv (Cat\{ Cat(\mathbf{F}^s_{low},\  \mathbf{E}^s_{low}),\   Cat(\mathbf{F}^q_{low},\  \mathbf{E}^q_{low})\}),
\end{equation}
where the superscript “$0$” of $\mathbf{F}^{0}_{low}$ represents the raw features. 

In FAM, the input low-level support and query features $\in \mathbb{R}^{C^{'} \times H^{'} \times W^{'}}$ first undergo a 3x3 convolution followed by GELU activation~\cite{hendrycks2023gaussian}. 
Next, the features pass through a 1x1 convolution and a max pooling operation to decrease the size of the features.
The convolution operations will capture local spatial features and non-linearity operations enhance the representation of features.
We obtain the intermediate features, which can be summarized as:
\begin{equation}
    \label{eq:6}
    MaxPool(Conv1(GELU(Conv3(\mathbf{F}^{0}_{low})))) \in \mathbb{R}^{\frac{C^{'}}{2} \times \frac{H^{'}}{2} \times \frac{W^{'}}{2}},
\end{equation}
where $Conv1$ and $Conv3$ are 1x1 and 3x3 convolution operations, respectively.
$MaxPool$ is a max pooling operation and $GELU$ denotes the GELU function.
Then, we apply the same feature processing steps again to obtain the final low-level features $\mathbf{F}_{low} \in \mathbb{R}^{C^{'} \times H^{'} \times W^{'}}$. Note that the dimensions are matched to the final high-level features $\mathbf{F}_{high}$.

\subsection{Loss Function}
There are two primary parts in our loss function:
\ding{182} We calculate pixel-wise cross-entropy loss and augment it with dice loss~\cite{deng2015semantic} for better performance, which calculates the intersection over the union between the predicted mask $\mathbf{M}_{q}$ and ground truth masks $\mathbf{T}_{q}$.
We define the loss function as the mask loss $\mathcal{L}_{mask}$.
\ding{183} Following~\cite{underwater2023iros}, we calculate the loss for aligning the predicted segmentation of support images with their ground truth, the align loss $\mathcal{L}_{align}$.

Overall, our final loss $\mathcal{L}$ will be:
\begin{equation}
    \label{eq:loss}
    \mathcal{L} = \mathcal{L}_{mask} + \mathcal{L}_{align}
\end{equation}
It is important to note that $\mathcal{L}_{align}$ is averaged over the five support images in the 5-shot setting.
\section{Experiments}
\subsection{Experimental Setup}
\noindent\textit{\textbf{Datasets.}}
We evaluate the performance of our FSSUWNet on two real-world few-shot semantic segmentation underwater image datasets, namely UWS~\cite{underwater2023iros} and SUIM-FSS based on SUIM dataset~\cite{islam2020suim}.
UWS is the animal-centric dataset consisting of 576 underwater images, each with detailed pixel-level annotations, including 21 underwater animals.
SUIM provides detailed pixel-level annotations for underwater semantic segmentation, consisting of 1635 images including eight semantic categories.
The cross-validation experiments are typically used in existing FSS works~\cite{lang2022bam,lang2023bam,apanet,underwater2023iros}.Following~\cite{underwater2023iros}, UWS dataset is evaluated with four-fold cross-validation, i.e., training our model on three of them while using the fourth for evaluation.

Given the limited segmentation categories in the SUIM dataset for four-fold validation, we divided it into two balanced groups, forming the SUIM-FSS dataset evaluated with two-fold cross-validation. After excluding the background (waterbody), SUIM contains seven target categories: Human divers (HD), Aquatic plants and sea-grass (PF), Wrecks and ruins (WR), Robots (RO), Reefs and invertebrates (RI), Fish and vertebrates (FV), and Sea-floor and rocks (SR). These categories were balanced and divided into two-fold datasets, as shown in Table~\ref{tab:suim-fss}. 
For ground truth extraction, we utilized the semantic segmentation masks in SUIM, removing instances where the target pixel count was less than 10\% of the total image pixels. During training, for each fold, we randomly sampled 1,000 pairs of support and query images from the same class. Test splits were drawn from unseen classes, and image pairs were maintained to ensure reproducibility of model evaluation.
\begin{table}[!t]
\centering
\caption{
    Details of the SUIM-FSS dataset based on SUIM dataset.
}
\label{tab:suim-fss}
    \begin{tabular}{l|cccc|ccc}
    \hline
    & \multicolumn{4}{c|}{\textbf{Split-0}} & \multicolumn{3}{c}{\textbf{Split-1}} \\
    \hline
    \textbf{Class} & HD & PF & RI & RO & FV & SR & WR \\
    \textbf{Number} & 142 & 117 & 160 & 99 & 160 & 160 & 160 \\ \hline
    \end{tabular}
\end{table}
%

\noindent\textit{\textbf{Evaluation Metric.}}
Following the settings of previous works~\cite{pfenet,apanet,lang2022bam,lang2023bam,underwater2023iros}, the mean Intersection over Union (mIoU) serves as the evaluation metric
for our experiments, which involve the IoU score and then computing the overall average of these scores.

\subsection{Implementation Details}
We initialize the Shared Feature Encoder with the VGG-16 pretrained model~\cite{vgg} and implement the Feature Enhanced Encoder using a simplified variant of Segformer~\cite{xie2021segformer}, which provides lightweight and efficient multi-scale feature extraction. As an additional component, SegFormer leverages its feature extraction strengths without significantly increasing computational complexity.
Following the settings in~\cite{underwater2023iros}, training was conducted for 40 epochs with a batch size of 1, using an SGD optimizer with a learning rate of 0.001, reduced by a factor of 0.1 every 10,000 iterations. Momentum and weight decay were set to 0.9 and 0.0005, respectively. Low-level features $\mathit{\mathbf{F}^i_{low}}$ were taken from the last layers of $conv2_x$, and high-level features $\mathit{\mathbf{F}^i_{high}}$ were obtained by concatenating the outputs from $conv4_x$ and $conv5_x$. All experiments were conducted on a single NVIDIA A100 GPU.

\begin{table*}
    \caption{Performance comparison on four folds of UWS dataset in terms of mIoU. The ``Mean'' row represents the averaged class mIoU across four folds. The best and second-best performances are in bold and underlined respectively.
    } 
    \centering
    \begin{center}
    \resizebox{0.75\linewidth}{!}{
        \begin{tabular}{ l |  c  c  c  c | c |  c  c  c  c | c  }
            \toprule
            \multirow{2}{*}{\textit{Methods}}  & \multicolumn{5}{c|}{1-Shot} & \multicolumn{5}{c}{5-Shot}   \\ 
            \cline{2-11}
            & Fold-0 & Fold-1 & Fold-2 & Fold-3 & Mean & Fold-0 & Fold-1 & Fold-2 & Fold-3 & Mean \\
            
            \specialrule{0em}{0pt}{1pt}
            \hline
            \specialrule{0em}{1pt}{0pt}
            
            PANet \cite{panet}
            & 69.11 & 59.92 & 65.76 & 67.29 & 65.52
            & 71.57 & 62.96 & 68.22 & 69.59 & 68.09 
             \\  
            PMMs \cite{PMMs2020}
            & 68.76 & \textbf{63.86} & 65.70 & 70.03 & 67.08 
            & 71.49 & 63.79 & 68.26 & 70.91 & 68.61
             \\   
            HSNet \cite{min2021hypercorrelation}
            & 62.81 & 57.39 & 61.38 & 63.09 & 61.17
            & 68.01 & 62.62 & 68.87 & 69.85 & 67.34
             \\     
            ASNet \cite{kang2022ifsl}
            & 62.93 & 58.25 & 67.04 & 64.42 & 63.16
            & 68.37 & 65.23 & \underline{71.91} & \underline{73.26} & \underline{69.69}
            \\
            BAM \cite{lang2023bam}
            & 69.42 & 59.66 & 57.39 & 57.85 & 61.08
            & 71.42 & 60.05 & 60.13 & 62.45 & 63.51
            \\
            UWSNetV2 \cite{underwater2023iros} 
            & 70.48 & 62.36 & 67.50 & \underline{70.22} & \underline{67.64}
            & 74.20 & 62.91 & 70.09 & 70.77 & 69.49
            \\
            UWSNetV6 \cite{underwater2023iros} 
            & \underline{70.66} & 61.75 & \underline{68.13} & 69.81 & 67.59
            & \underline{74.21} & \underline{63.96} & 70.08 & 70.48 & 69.68
            \\             
            \specialrule{0em}{0pt}{1pt}
            \hline
            \specialrule{0em}{1pt}{0pt}
            \rowcolor[gray]{0.95} FSSUWNet (ours)
            & \textbf{73.23} & \underline{63.62} & \textbf{69.25} & \textbf{72.11} & \textbf{69.55} 
            & \textbf{74.92} & \textbf{65.62} & \textbf{72.06} & \textbf{73.36} & \textbf{71.49}
            \\            
            \bottomrule            
        \end{tabular}
   }
    \end{center}
    \label{tab:compare_uws}
\end{table*}

\subsection{Experimental Results}
\noindent\textit{\textbf{Quantitative Results.}}
The comparative results of our approach on the UWS and SUIM-FSS datasets are presented in Table~\ref{tab:compare_uws} and Table~\ref{tab:compare_suim}, with the mean Intersection over Union (mIoU) metric used to evaluate performance.
It is evident that our FSSUWNet significantly outperforms advanced FSS methods across all settings, achieving new state-of-the-art.
The proposed method surpasses the previous best (UWSNetV2~\cite{underwater2023iros} and ASNet~\cite{kang2022ifsl}) and by 2.8\% and 2.6\% in the mIoU metric for 1-shot and 5-shot scenarios in UWS dataset, respectively.
Additionally, our method is the only one to exceed 70\% mIoU in the 5-shot setting, achieving a score of 71.49.
In Table~\ref{tab:suim-fss}, we use the model with only the Shared Feature Encoder as the baseline for our method.
We can see that in the SUIM-FSS dataset, which includes collaborative scenarios between robots and divers, our approach achieves the top performance, surpassing UWSNetV2 by 3.32\% in the 1-shot setting.

\noindent\textit{\textbf{Qualitative Results.}\label{label:visual}}
\begin{figure*}[!t]
\centering
\includegraphics[width=0.8\textwidth]{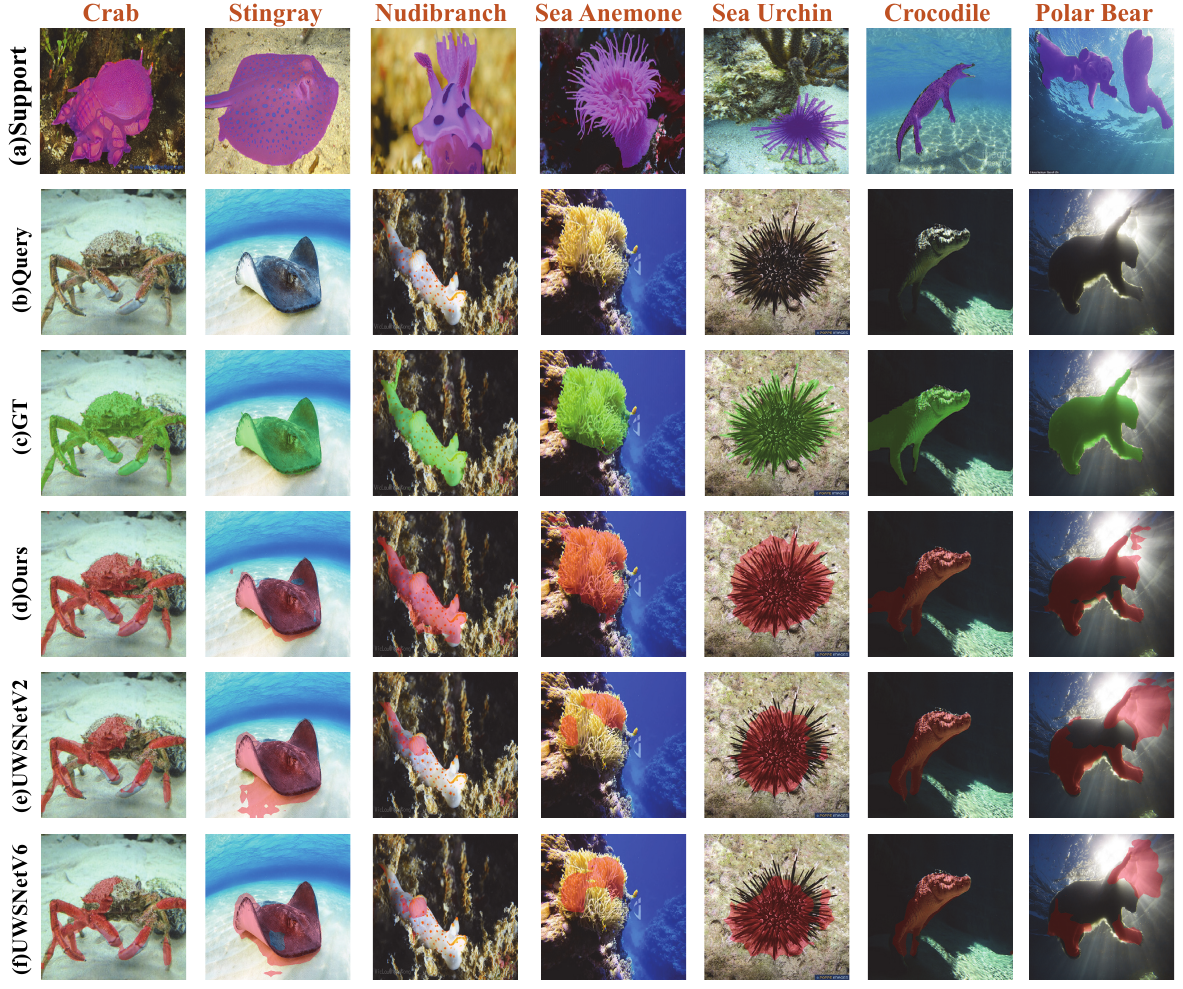}
\caption{Qualitative results of the proposed FSSUWNet and other methods on UWS dataset. From top to bottom: (a) support images with ground-truth masks, (b) query images, (c) query images with ground-truth masks, (d) results of FSSUWNet and (e) results of UWSNetV2, and (f) results of UWSNetV6.}
\label{fig:uws}
\end{figure*}

\begin{table}
    \caption{Performance comparison on SUIM dataset in terms of mIoU. The ``Mean'' row represents the averaged class mIoU across seven classes. The best results are highlighted in bold.} 
    \centering
    \resizebox{0.5\textwidth}{!}{
        \begin{tabular}{ l |  c  c  | c | c  c | c }
            \toprule
            \multirow{2}{*}{\textit{Models}}  & \multicolumn{3}{c|}{1-Shot} & \multicolumn{3}{c}{5-Shot}   \\ 
            \specialrule{0em}{0pt}{1pt}
            \cline{2-7}
            \specialrule{0em}{1pt}{0pt} 
            & Fold-0 & Fold-1 & Mean & Fold-0 & Fold-1 & Mean \\
            \specialrule{0em}{0pt}{1pt}
            \hline
            \specialrule{0em}{1pt}{0pt}
            Baseline
            & 39.12 & 39.69 & 39.41 & 44.76 & 45.88 & 45.32 \\
            \specialrule{0em}{0pt}{1pt}
            \hline
            \specialrule{0em}{1pt}{0pt}
            UWSNetV2 \cite{underwater2023iros} 
            & 38.09 & 40.71 & 39.40 & 45.63 & 52.53 & 49.08 \\
            \specialrule{0em}{0pt}{1pt}
            \hline
            \specialrule{0em}{1pt}{0pt}
            UWSNetV6 \cite{underwater2023iros} 
            & 40.13 & 41.08 & 40.61 & 44.49 & 52.42 & 48.46\\
            \specialrule{0em}{0pt}{1pt}
            \hline
            \specialrule{0em}{1pt}{0pt}
            \gt FSSUWNet (ours)
            & \textbf{41.40} & \textbf{44.03} & \textbf{42.72} 
            & \textbf{46.04} & \textbf{52.81} & \textbf{49.43}\\
            \bottomrule            
        \end{tabular}
    }
    \label{tab:compare_suim}
\end{table}
To illustrate the effectiveness of our proposed model,
we further visualize the segmentation results and compare them with those of other methods, as shown in Figure~\ref{fig:uws} for the UWSNet dataset.
It can be found that: For UWSNet, the third and fourth columns in Figure~\ref{fig:uws} indicate that only our method successfully predicts the target object in the query image, while other methods fail.
Besides, in the fourth column (sea anemone), the target object in the support image does not resemble the one in the query image, yet our method still predicts the segmentation mask accurately.
The sixth column (crocodile) represents a common low-light scenario in underwater images. Compared to other methods, our method can successfully predict the lower left half of the crocodile in the dark area. 
Additionally, the seventh column (polar bear) illustrates a scenario with complex underwater lighting, causing other methods to mistakenly identify background areas as foreground, while our method still achieves perfect segmentation results.
These visual results demonstrate the efficacy of our proposed Feature Enhanced Encoder.
 
\subsection{Ablation Study and Analysis}
\begin{table*}
    \caption{Ablation Study on UWS dataset in terms of mIoU. The ``Mean'' row represents the averaged class mIoU across four folds. The results of the complete model are highlighted in bold.} 
    \centering
    \resizebox{0.8\textwidth}{!}{
        \begin{tabular}{ l |  c  c  c  c | c |  c  c  c  c | c  }
            \toprule
            \multirow{2}{*}{\textit{Methods}}  & \multicolumn{5}{c|}{1-Shot} & \multicolumn{5}{c}{5-Shot}   \\ 
            \specialrule{0em}{0pt}{1pt}
            \cline{2-11}
            \specialrule{0em}{1pt}{0pt} 
            & Fold-0 & Fold-1 & Fold-2 & Fold-3 & Mean & Fold-0 & Fold-1 & Fold-2 & Fold-3 & Mean \\
            
            \specialrule{0em}{0pt}{1pt}
            \hline
            \specialrule{0em}{1pt}{0pt}
            Baseline (w/o FEE)
            & 70.75 & 60.59 & 68.28 & 70.99 & 67.65 \rt{(+1.90)}
            & 73.83 & 63.84 & 70.82 & 72.00 & 70.12 \rt{(+1.37)}
            \\
            \specialrule{0em}{0pt}{1pt}
            \hline
            \specialrule{0em}{1pt}{0pt}
            w/o FAM
            & 71.42 & 63.18 & 68.22 & 70.05 & 68.22 \rt{(+1.33)}
            & 74.14 & 65.46 & 70.26 & 71.91 & 70.44 \rt{(+1.05)}
            \\ 
            \specialrule{0em}{0pt}{1pt}
            \hline
            \specialrule{0em}{1pt}{0pt}
            w/ $\mathit{\mathbf{F}_{high}}$ $\leftrightarrow$ $\mathit{\mathbf{F}_{low}}$
            & 43.72 & 41.33 & 38.79 & 39.34 & 40.65 \rt{(+28.90)}
            & 47.76 & 47.22 & 47.16 & 45.21 & 46.84 \rt{(+24.65)}
            \\   
            \specialrule{0em}{0pt}{1pt}
            \hline
            \specialrule{0em}{1pt}{0pt}
            \rowcolor[gray]{0.95} FSSUWNet
            & \textbf{73.23} & \textbf{63.62} & \textbf{69.25} & \textbf{72.11} & \textbf{69.55} 
            & \textbf{74.92} & \textbf{65.62} & \textbf{72.06} & \textbf{73.36} & \textbf{71.49}
            \\            
            \bottomrule            
        \end{tabular}
    }
    \label{tab:ablation}
\end{table*}
To better analyze and understand the proposed FSSUWNet, we conducted a series of ablation experiments on the UWS dataset, as shown in Table~\ref{tab:ablation}. 

\subsubsection{Ablation Study on FEE}
FEE is used to extract features that complement those from the pre-trained model and is one of the core components of our method. 
As shown in the first row of Table~\ref{tab:ablation}, ``Baseline'' refers to the model where FEE is removed, i.e., FSSUWNet without FEE.
Compared to FSSUWNet, FEE brings performance improvements across four-fold validation in terms of the mIoU scores, with a 1.90\% increase in the 1-shot setting.
In Table~\ref{tab:suim-fss}, we compare the performance of the baseline and FSSUWNet on the SUIM-FSS dataset. The model with FEE (FSSUWNet) outperforms the model without FEE (Baseline) by 3.31\% in 1-shot and 4.11\% in 5-shot settings.
Both qualitative and quantitative results indicate that our proposed FEE can improve the model's performance in processing underwater images compared to the baseline.
%
\subsubsection{Ablation Study on FAM}
To analyze the impact of the Feature Alignment Module (FAM) used to align low-level features $\mathbf{L}^0_{low}$ with high-level features, we designed an experiment by removing FAM from the model, leaving only the raw low-level features and high-level semantic features.
As illustrated in Table~\ref{tab:ablation}, the results of the second row indicate that without FAM leads to a decline in model performance compared to the FSSUWNet which includes the low-level features that have been enhanced, i.e. $\mathbf{F}_{low}$.
Note that $\mathbf{F}^0_{low}$ is the raw features without FAM while $\mathbf{F}_{low}$ is the features operated by FAM.
It is suggested that $\mathbf{F}_{low}$ are beneficial for the FSS task and the alignment with high-level features is useful.
We attribute this benefit to the ability of low-level features to represent fundamental characteristics of the underwater images, such as the significant reduction in the red channel.
Our FAM further enhances and consolidates the capability of $\mathbf{F}^0_{low}$.
The information brought by FAM can be considered as global prior knowledge for segmentation and target recognition.

\subsubsection{Ablation Study on Features from Different Levels\label{levels}}
As discussed before, different levels of features play distinct roles in the network. To further validate our conclusions regarding the impact of low-level and high-level features in our method, we designed experiments to explore the differences between these features.
A direct idea is to interchange the roles of low-level and high-level features in FSSUWNet. Specifically, the high-level features ($\mathbf{H}_{s}$ and $\mathbf{H}_{q}$) output by the encoders are fed into the FAM as low-level features $\mathbf{F}^0_{low}$, while the low-level features ($\mathbf{L}_{s}$ and $\mathbf{L}_{q}$) are treated as high-level features $\mathbf{F}_{high}$. 
It is important to note that, due to the difference in size between these feature types, we follow the same technology as FSSUWNet, i.e., adjusting the size of features from FAM to match that of the high-level features.
The results of swapping low-level and high-level features are shown in Table~\ref{tab:ablation} under ``$\mathit{\mathbf{F}_{high}}$ $\leftrightarrow$ $\mathit{\mathbf{F}_{low}}$''. It can be observed that after the exchange, the model's performance drops significantly, with decreases of 28.90\% and 24.65\% for 1-shot and 5-shot scenarios, respectively. 
This indicates that the new feature handling method is unsuitable for the FSS task in underwater images compared to the original approach.
In conclusion, the experimental results further validate the correctness of our approach to utilizing low-level features and the necessity of the proposed FAM. Additionally, we believe this conclusion can inspire the design of model architectures in future related underwater image FSS work.

\section{Conclusion And Future Works}
In this paper, we show that pre-trained models in current FSS frameworks, e.g., VGG-16, could be fragile when encountering underwater images.
To address the challenge, we propose a novel FSS framework tailored for underwater images by leveraging the complementary features extracted from an auxiliary encoder and the different levels of image features, namely FSSUWNet.
Additionally, we propose SUIM-FSS, a cross-validation dataset version based on the SUIM underwater image dataset.
Extensive experiments on underwater image segmentation datasets and network module ablation experiments validate the effectiveness and adaptability of our proposed FSSUWNet.
In the future, we will deploy our approach on underwater devices, e.g., remote-operated vehicles, for real-world marine experiments, further validating the performance of current FSS algorithms in underwater application scenarios.

\bibliographystyle{IEEEtran}
\bibliography{ref}

\end{document}